\documentclass{article}

\PassOptionsToPackage{numbers, compress}{natbib}
\usepackage[final]{nips_2017}

\usepackage[utf8]{inputenc} 
\usepackage[T1]{fontenc}    
\usepackage{hyperref}       
\usepackage{url}            
\usepackage{booktabs}       
\usepackage{amsfonts}       
\usepackage{nicefrac}       
\usepackage{microtype}      
\usepackage{graphicx}
\usepackage{amsmath}
\usepackage{amssymb}
\usepackage{amsthm}
\usepackage{xcolor}
\usepackage{soul}
\usepackage[caption=false, font=footnotesize]{subfig}

\newcommand{\william}[1]{{\footnotesize \sf {\color{blue}{\bf William:} #1}}}
\newcommand{\brad}[1]{{\footnotesize \sf {\color{red}{\bf Brad:} #1}}}

    \renewcommand{\william}[1]{}
    \renewcommand{\brad}[1]{}

\title{Context-modulation of hippocampal dynamics and deep convolutional networks} 

\author{
  James B.~Aimone \\
  Center for Computing Research\\
  Sandia National Laboratories\\
  Albuquerque, NM 87185-1327 \\
  \texttt{jbaimon@sandia.gov} \\
  \And
  William M.~Severa \\
  Center for Computing Research\\
  Sandia National Laboratories\\
  Albuquerque, NM 87185-1327 \\
  \texttt{wmsever@sandia.gov} \\
}

\begin{document}

\maketitle

\begin{abstract}
	Complex architectures of biological neural circuits, such as parallel processing pathways, has been behaviorally implicated in many cognitive studies.  However, the theoretical consequences of circuit complexity on neural computation have only been explored in limited cases. Here, we introduce a mechanism by which direct and indirect pathways from cortex to the CA3 region of the hippocampus can balance both contextual gating of memory formation and driving network activity.  
	 We implement this concept in a deep artificial neural network by enabling a context-sensitive bias.  The motivation for this is to improve performance of a size-constrained network.  Using direct knowledge of the superclass information in the CIFAR-100 and Fashion-MNIST datasets, we show a dramatic increase in performance without an increase in network size. 
\end{abstract}

\section{Introduction}
Although much of the attention on leveraging biological complexity in machine learning has focused on spiking opposed to non-spiking neurons, the difference in circuit architecture complexity between artificial neural networks (ANNs) and neural circuits is equally striking.  ANNs are typically feedforward or leverage local recurrence; however, outside of early sensory processing, most biological neural circuits are far more complex.  While the Felleman and van Essen~\cite{felleman1991} circuit for the visual system may be the most famous, the brain is flush with examples of parallel pathways, extended loops, and consolidation of multiple inputs.  In this short paper, we explore how parallel pathways could be useful for conveying information at multiple scales, in effect providing an efficient mechanism for conveying global context and local meaning simultaneously.  

To examine the effect of multiple inputs on a neuron, specifically the unique case where a neural population receives both a direct and an indirect input from a common source, we specifically look at the effects of convergent population inputs onto neurons within the CA3 of the hippocampus, in which neurons receive direct input from the cortex and an indirect cortical input that passes through the dentate gyrus (DG). While the hippocampus is more commonly associated with memory formation at the circuit level, we consider it here simply for its well characterized and straightforward local structure.  From this anatomy, we infer a unique function of separated synaptic inputs as it relates to conveying specific and coarse information.  




\section{Context modulation of Hippocampal neurons through dual pathways}
\begin{figure}[h]
  \centering
  \subfloat[]{\includegraphics[width=.3\textwidth]{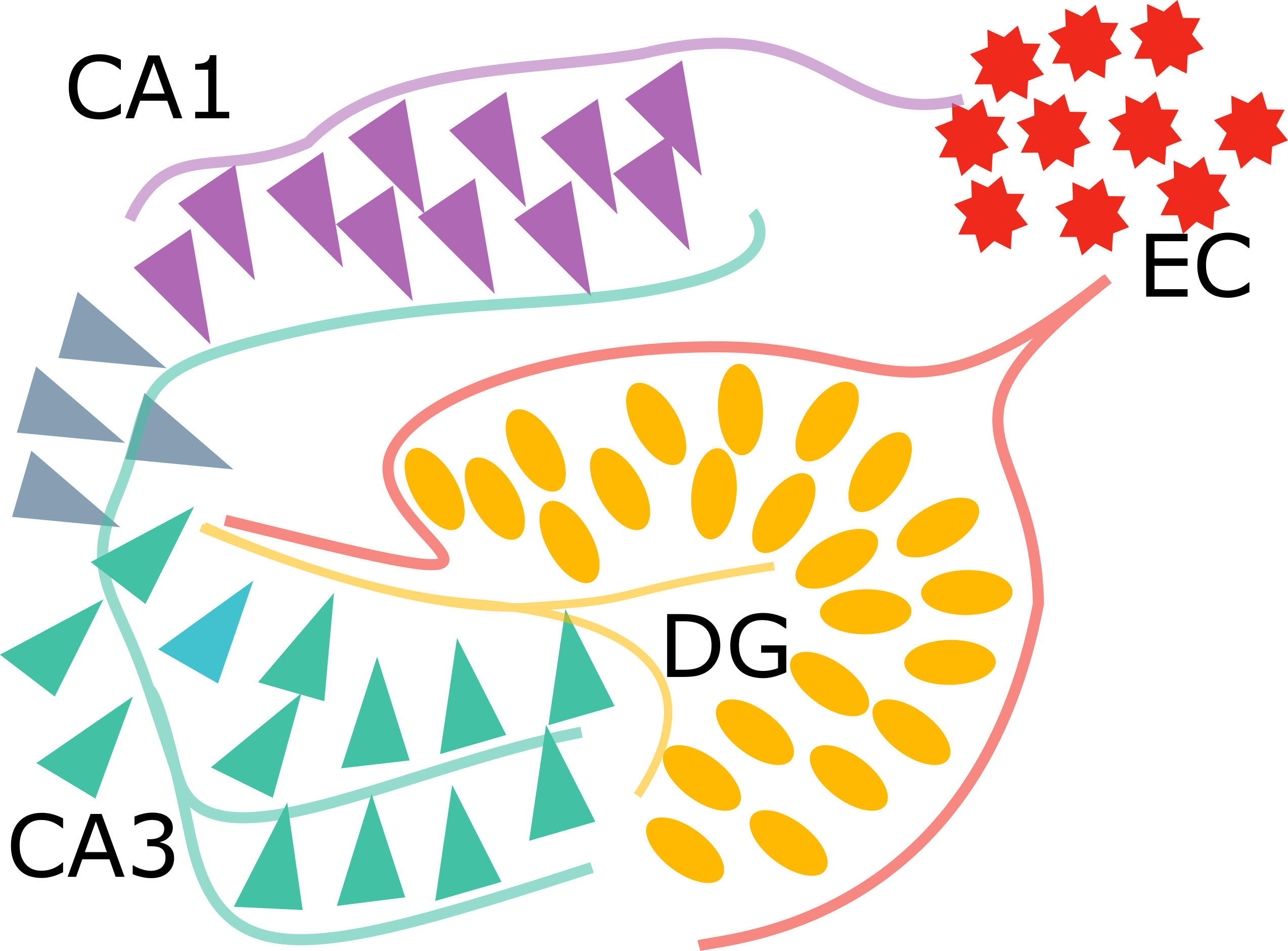} \label{hippoFigA}
}
\subfloat[]{  \includegraphics[width=.2\textwidth]{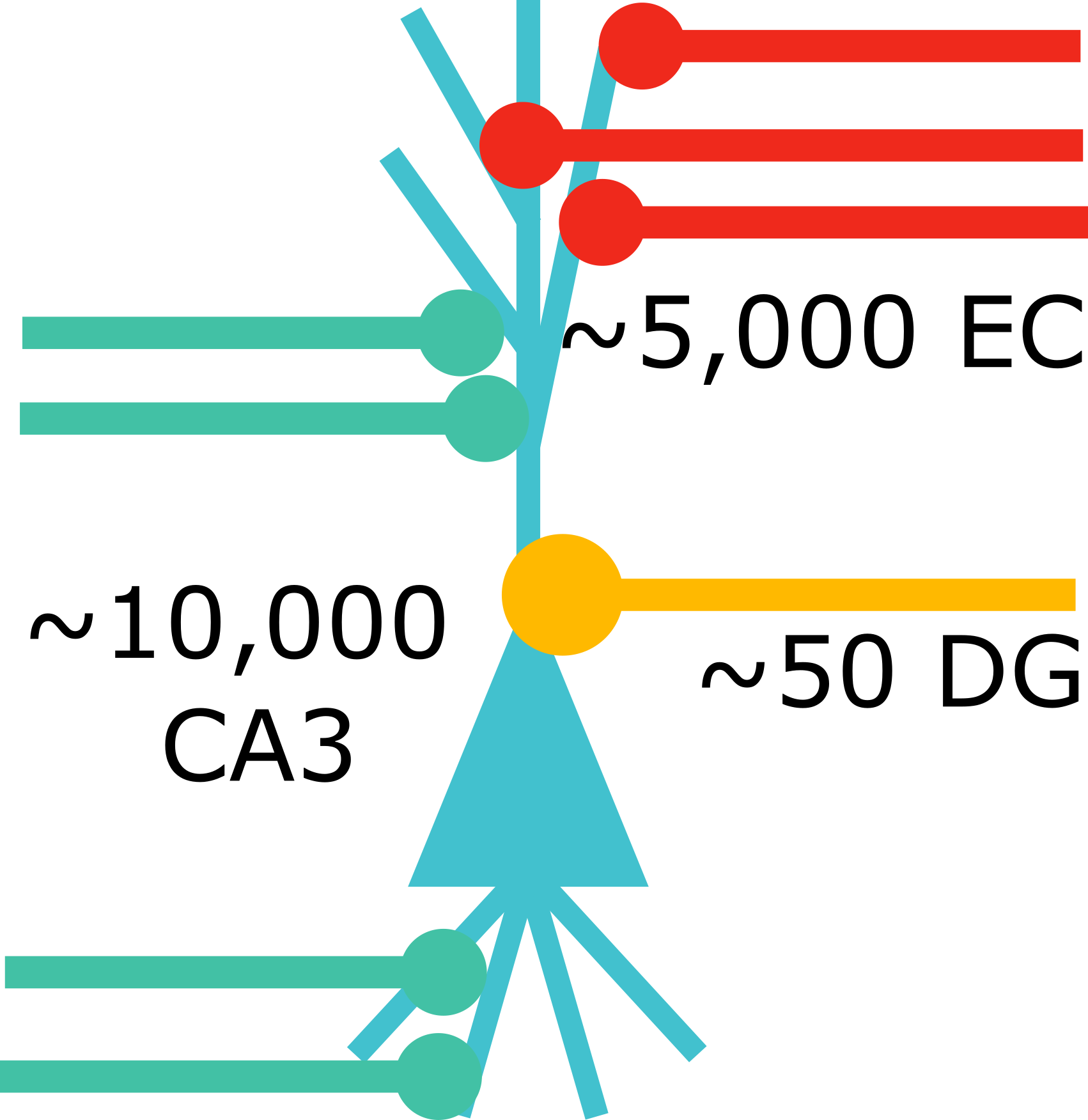}\label{hippoFigB}
}\\
\subfloat[]{\includegraphics[width=.55\textwidth]{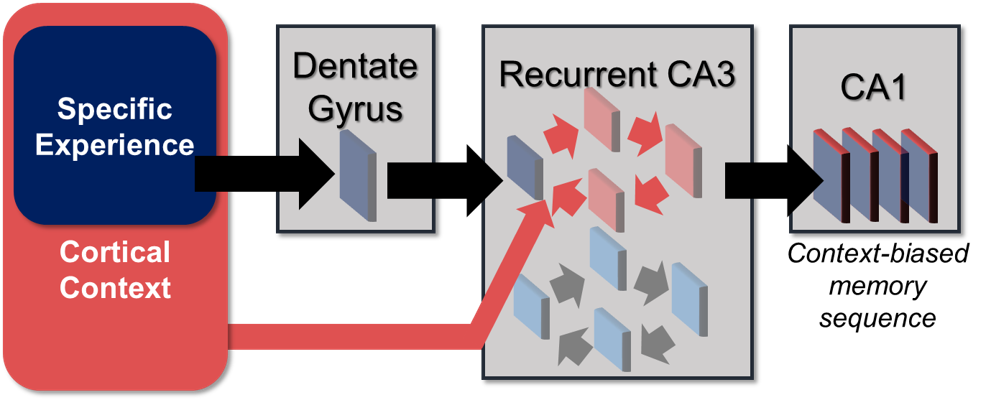} \label{hippocampusDiagram}
}
  \caption{(a) Illustration of subregions of hippocampus circuit.  (b) Detailed view of excitatory inputs onto CA3 pyramidal neuron.  (c) Schematic of direct cortical inputs providing context bias for processing in the hippocampus. \william{Subfitures are not labelled.  Maybe combine this figure with the hipp schematic below and then the dl schematic with the results plot?}}

\end{figure}

First, to illustrate how the brain potentially uses parallel processing pathways, we consider a particular case within the hippocampus.  While the hippocampus shares few similarities to modern artificial networks, it has a clear anatomical structure that makes it amenable to analysis, consisting of three primary sub-regions (the DG, CA3, and CA1) that each receive cortical input from the entorhinal cortex (EC) and a well-understood connectivity (Figure~\ref{hippoFigA}).  


Here, we focus on the influence of the direct (EC) projection and the indirect (EC via DG) projection on the short-term dynamics of the CA3 network.  We ignore for now the extensive forms of plasticity, including lifelong neurogenesis in the DG~\cite{aimone2011resolving}, recurrent synaptic plasticity in CA3~\cite{rebola2017operation} and robust synaptic turnover in CA1~\cite{attardo2015impermanence}.  While the DG does receive some inputs aside from EC, for the most part the DG is thought to recode EC inputs, \textbf{suggesting that the CA3 receives two versions of the same information}, anatomically flanking the numerous recurrent connections from within the CA3 network (Figure~\ref{hippoFigB}). The conjunction of two anatomically distinct yet informatically overlapping projections onto the recurrent CA3 network has been key to many influencial theoretical studies of hippocampus function~\cite{treves1992computational}, however here we consider whether this paradigm offers a general capability to artificial networks.

First, we look at the anatomical structure of these inputs (Figure~\ref{hippoFigB}).  While there are only a few DG inputs on a CA3 pyramidal neuron, they are massive --- often covering several dendritic spines.  The proximity and strength of these synapses suggests that when they are active, they can ``detonate'' the downstream neuron highly efficiently~\cite{Henze2002}. On the other hand, the EC inputs, while numerous, are on the distal dendrites far away from the soma and, unlike the DG inputs, likely provide only a moderate current to the neuron.  In between lie the recurrent connections, which are both numerous and efficacious, consistent with other models of hippocampal and cortical recurrent dynamics.

Second, we consider the nature of the external inputs' representations.  DG neurons are known to be very sparsely active---with only a small percentage active at a given time and highly decorrelated even within a spatial context\cite{severa2016combinatorial}.  Thus given the sparse input numbers, we should expect that generally the DG inputs are silent, interrupted by occasional bursts from a single DG neuron. In contrast, the EC inputs have a more distributed code, either as grid cells encoding space or other high level cortical representations~\cite{fiete2008grid}.  Thus across the approximately 5,000 synaptic inputs as a whole, we should expect a fairly smooth transitions in combined input activity, particularly within a given environment.  

Combined, these two observations suggest the model of CA3 neuron modulation shown in Figure~\ref{hippocampusDiagram}.  While the dynamics of the CA3 network are likely dominated by the numerous and reasonably powerful recurrent connections, the cortex can use these two distinct inputs to control these dynamics.  For one, the sparse and powerful indirect inputs through the DG are well-suited to \textit{instantaneously} set the recurrent network in a desired state---a highly distinct state, per the DG's presumed pattern separation function---which then quickly evolves per the recurrent connections.  At the same time, the slowly evolving and diffuse direct inputs are ideal for setting the \textit{context} of these dynamics.  These weaker inputs are not necessarily able to drive activity directly, but they can bias the recurrent activity in meaningful ways that can similarly influence computation, albeit at far broader time scales.

One can hypothesize several benefits for having context reflected this way in hippocampal memory encoding (Figure~\ref{hippocampusDiagram}).  First, having relatively small subsets of neurons pre-selected by the EC would be well-suited for fast \mbox{``one-shot''-like} learning that minimizes interference with other learned representations.  Such rapid learning may either occur within the recurrent CA3 connections~\cite{rebola2017operation} or may be better suited for the feed-forward Schaffer Collateral connection to CA1.  Indeed, recent results showing rapid CA1 synaptic restructuring would be consistent with that region continually realigning itself to these relevant CA3 ensembles~\cite{attardo2015impermanence}.  At the same time, as these context-selected subsets of neurons do not need to be independent, there are effectively a very large number of potential combinations available for any encoding.  Thus while the memory capacity within the recurrent CA3 network may be relatively low for any particular context, the overall capacity may be quite high.

\section{Impact of parallel pathways on context modulation of ANNs}
\william{\st{The analysis above demonstrates that the dual direct and indirect pathways from EC to CA3 can have a potent function in the context of hippocampus.  However, the broader applicability of parallel pathways as a computing mechanism, and thus their general applicability to ANNs, is worth exploring.}}  Some ANNs, such as deep residual networks, already leverage parallel pathways to facilitate learning~\cite{he2016deep}, although this use of parallel pathways was not necessarily biologically inspired. So we next sought to identify whether parallel inputs could provide context inputs that help constrain a network's function in a manner comparable to how we hypothesized hippocampus function may be improved. 






One key motivation for this development is the application of deep convolutional networks (DCNs) in memory-constrained environments.  These include the growing segment of embedded processors available for internet of things (IoT) devices as well as drones and autonomous vehicles.  
One promising direction is neural network compression, however, the neuroscience inspiration explained above may enable more efficient models.  The impact of context in the above CA3 study suggests that the correlation of EC input against a baseline sensitivity provides a bias of sorts to the overall CA3 dynamics.  Setting aside the specific downstream dynamics of CA3, the consideration of context in the biological setting is not altogether different from the concept of biases in ANNs.  In our thought experiment above, the net effect of shifting the EC input onto pyramidal neurons is comparable to shifting the threshold required for other inputs to drive it to fire (Figure~\ref{DCNContextFig}).



We explored our model of context switching on a simple DCN implementation.  Inspired the EC ``bias'' above, our ANN model adds a parameter and expands bias to be context-dependent.  For each input vector/image $x$, we assume there exist an accompanying context $\hat{x}$ which we represent with a one-hot vector encoding. 
Formally, the output of a traditional ANN layer is computed as $o(x) = f(Ax + b)$ where $A$ is a weight matrix, $b$ is the bias and $f$ is the activation function.  
For a context $\hat{x}$, instead of the classic bias, we use $o(x) = f(Ax + b_{\hat{x}})$ which represents the concept that the bias may depend on $\hat{x}$. Let $\delta_i = 1$ if the  context $\hat{x}$ is $i$ and $\delta_i = 0 $ otherwise.  In this case, our layer can equivalently be written
$o(x) = f(Ax + \sum_i \delta_i b_{\hat{x}}) = f(Ax + B\hat{x})$ where $B$ is a matrix representing each of the biases under each context.

\subsection{Using Context to Improve Classification}
\begin{figure}[ht!]
  \centering
  \subfloat[]{  \includegraphics[width=.67\textwidth]{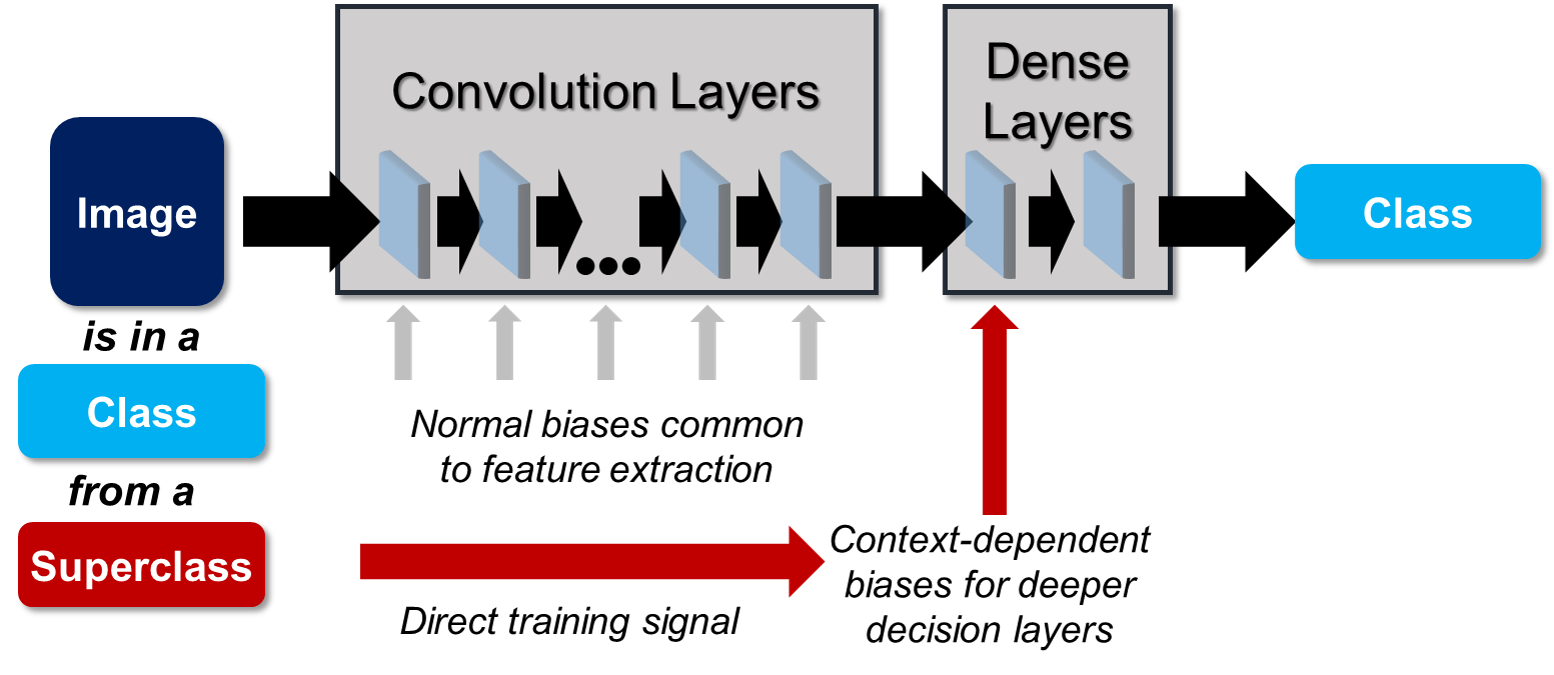}\label{DCNContextFig}
}\\
  \subfloat[]{ \includegraphics[height=2.5in]{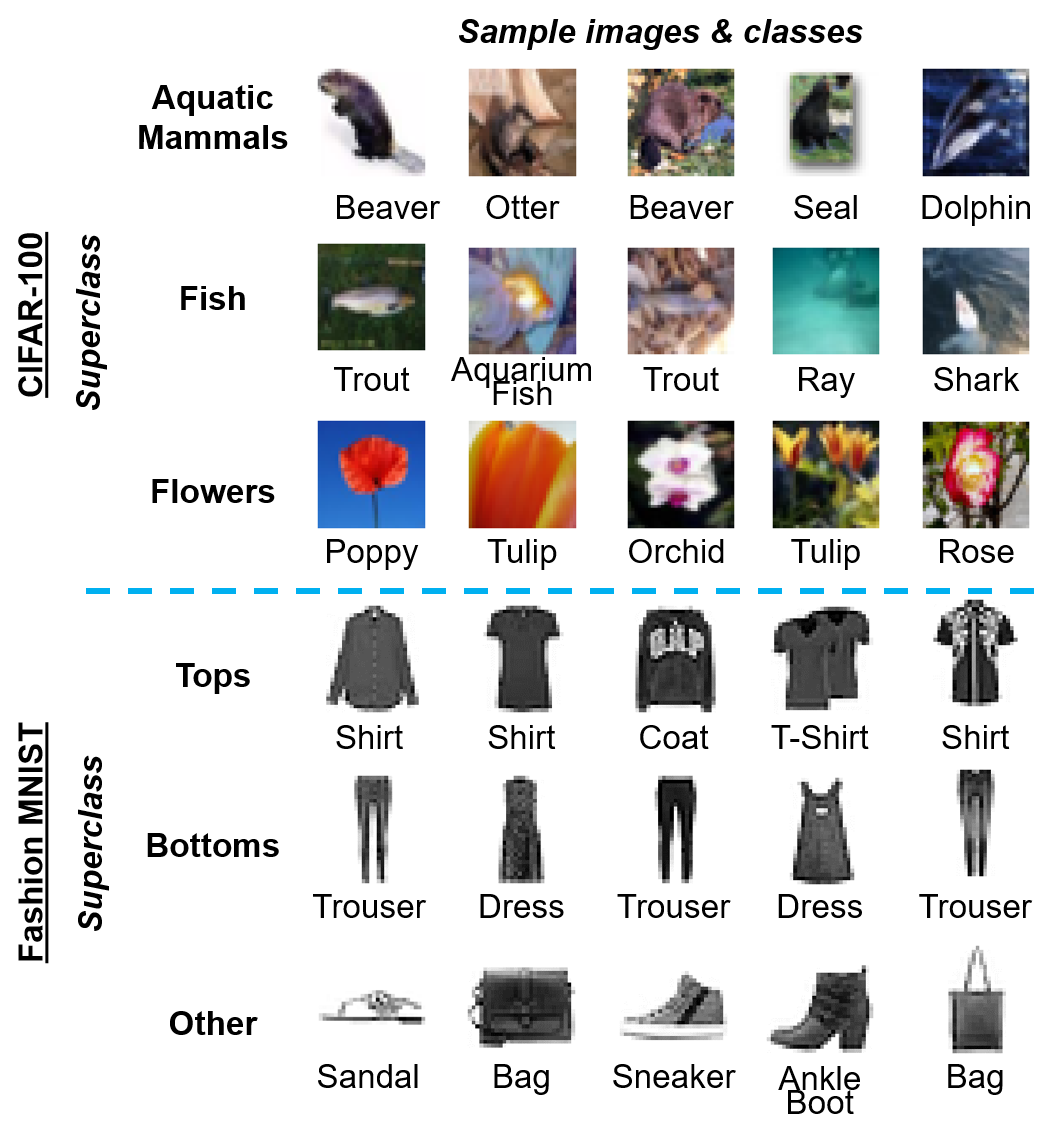}
    \label{fig:superclass}
}
  \subfloat[]{ \includegraphics[height=2.5in]{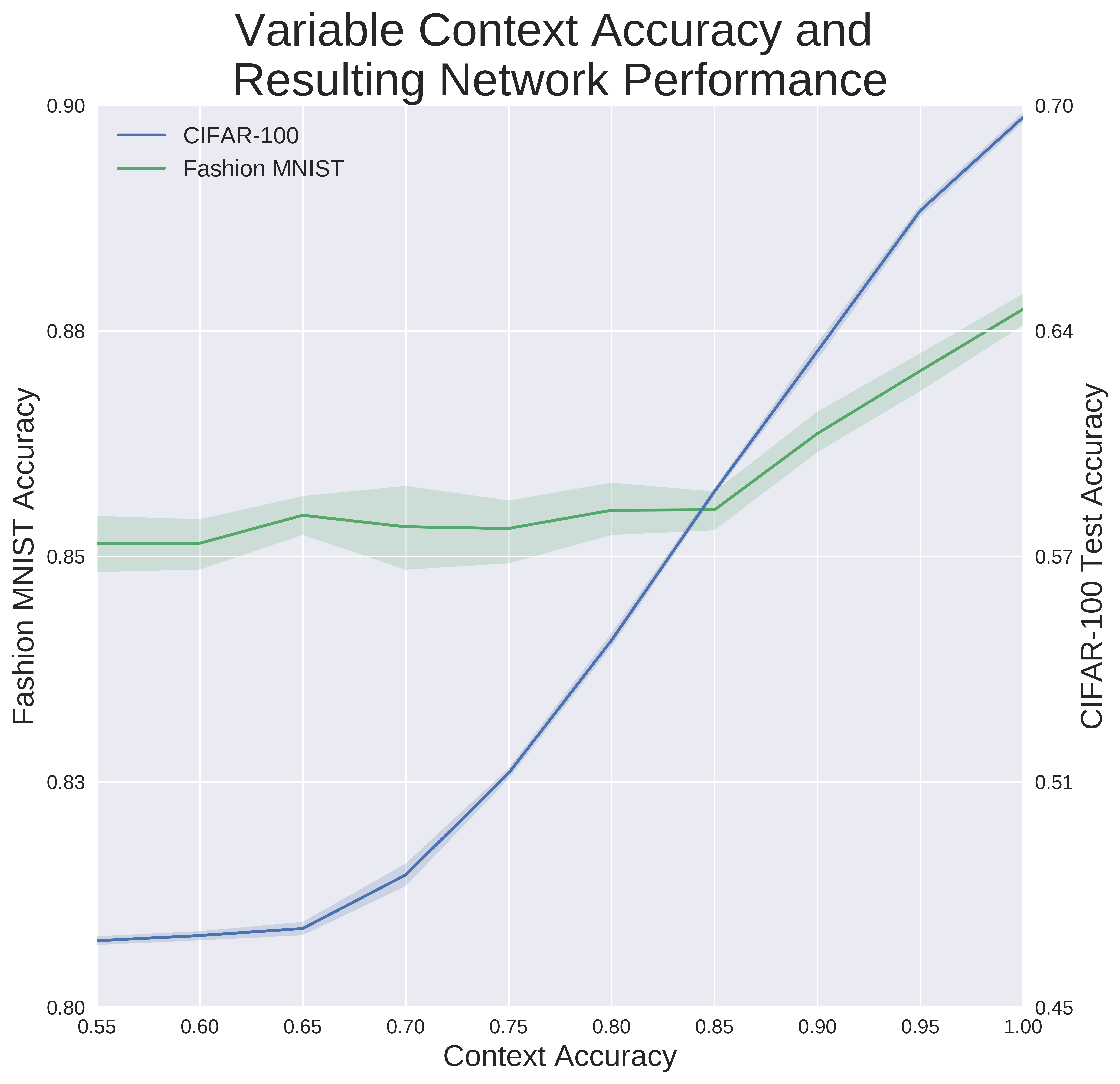}
    \label{fig:DCNResults}
    }
    
  \caption{(a) Schematic of direct superclass inputs providing context as bias inputs for VGG image processing.  (b) Examples of how different classes fall into superclasses within CIFAR and Fashion-MNIST. (c) Results from adding context to VGG Networks on CIFAR-100 and Fashion-MNIST.  The means over $10$ trials is plotted with the shaded areas representing $95\%$ confidence intervals.}

\end{figure}

We illustrate this model on the CIFAR-100 dataset~\cite{krizhevsky2009learning} and Fashion-MNIST~\cite{xiao2017fashion}(Figure \ref{fig:superclass}).  In CIFAR-100, each image belongs to both a superclass ($20$, also called a `coarse' class) and a standard class ($100$).  For Fashion-MNIST, we created coarse superclasses `Tops' (consisting of `T-shirt/top', `Pullover', `Coat', `Shirt'), `Bottoms' (`Trouser', `Dress') and `Other' (`Sandal', `Sneaker', `Bag', `Ankle boot').\william{or we can convert the previous sentence to a table to put next to the next figure?} We aim to classify the standard class by using the superclass as the context $\hat{x}$.  We use VGG16 pre-trained on ImageNet for feature extraction~\cite{simonyan2014very}.  After the convolutional layers, we employ two densely connected layers ($256$ exponential linear unit nodes, followed by a softmax layer).  We introduce the context into these layers at just the first dense layer.  Layers that received context used no bias.  We use dropout ($0.5$), to mitigate overfitting, and the adadelta optimizer~\cite{zeiler2012adadelta}. Tests were performed at a variety of context accuracy levels, wherein the superclass information was systematically degraded to simulate various superclass classifier performance.

The inclusion of superclass information improves the classification performance of our networks (Figure \ref{fig:DCNResults}).  As the quality of the superclass labels decrease, the benefit of context is diminished.  For Fashion-MNIST, any noise $>15\%$ in the context input offsets the benefit of the additional context.  However, context aids CIFAR performance even with noisy contextual input.  While the relative information content of the superclass labels likely explains some of the difference in context efficacy between these two data sets, future work is needed to help discern the differences seen here.

\section{Conclusions}
Even though hippocampal role in memory is quite a different function to image processing by DCNs, the results from both studies highlights the potential importance of parallel information pathways in neural computation.  The use of multiple pathways is commonly seen within neural circuits of the brain.  It is not unreasonable that the function of dual inputs proposed for the hippocampus here --- coarse information influences context while fine-grained information drives circuit behavior acutely --- may be general to many cortical and cortex-affiliated regions.  

Accordingly, the results here show that in artificial visual processing we can observe  that a parallel introduction of coarse and fine-grained information can improve performance markedly.  In our case providing superclass information to help guide the classification of images has a dramatic benefit on overall network performance.  It is not surprising that additional information helps classification, however there is no reason that the approach shown here cannot be extended to have the full system trained to learn both levels of information.  Further, as examples of adversarial limitations of deep networks continue to be introduced, it is worth considering that networks using parallel processing pathways to perform both coarse and fine classification may be more robust to dramatic sub-perceptual manipulation of low-level features.

\subsubsection*{Acknowledgments}

This work was funded by the DOE Advanced Simulation and Computing program and Laboratory-Directed Research and Development Program at Sandia National Laboratories.  Sandia National Laboratories is a multiprogram laboratory managed and operated by National Technology and Engineering Solutions of Sandia, LLC, for the U.S. Department of Energy’s National Nuclear Security Administration under contract DE-NA0003525.


\medskip

\small

\bibliographystyle{plainnat}
\bibliography{Aimone_CA3}

\end{document}